\pdfoutput=1
\documentclass[journal,twoside,web]{ieeecolor}
\usepackage[usenames,dvipsnames]{xcolor}
\usepackage{generic}
\usepackage{cite}
\usepackage{amsmath,amssymb,amsfonts}
\usepackage{algorithmic}
\usepackage{graphicx}
\usepackage{algorithm,algorithmic}
\usepackage{hyperref}
\hypersetup{hidelinks=true}
\usepackage{textcomp}
\def\BibTeX{{\rm B\kern-.05em{\sc i\kern-.025em b}\kern-.08em
    T\kern-.1667em\lower.7ex\hbox{E}\kern-.125emX}}
\usepackage{bbding}
\usepackage{booktabs}
\usepackage{multirow}
\usepackage{txfonts}
\usepackage{times}
\usepackage{latexsym}
\usepackage{enumerate}
\usepackage[T1]{fontenc}
\usepackage{amsmath}
\usepackage[utf8]{inputenc}
\usepackage{arydshln}
\usepackage{microtype}
\usepackage{inconsolata}
\usepackage{graphicx}
\usepackage{adjustbox}

\usepackage{pgfplots}
\usepackage{tikz}
\pgfplotsset{
   compat=1.17,
   legend entry/.initial=,
   every axis plot post/.code={%
       \pgfkeysgetvalue{/pgfplots/legend entry}\tempValue
       \ifx\tempValue\empty
           \pgfkeysalso{/pgfplots/forget plot}%
       \else
           \expandafter\addlegendentry\expandafter{\tempValue}%
       \fi
   },
}

\definecolor{blue}{HTML}{4F81BD}
\definecolor{red}{HTML}{C0504D}
\definecolor{green}{HTML}{9BBB59}
\definecolor{purple}{HTML}{9F4C7C}
\title{Automated Clinical Coding\\for Outpatient Departments}
\author{Viktor Schlegel, Abhinav Ramesh Kashyap, Thanh-Tung Nguyen, Tsung-Han Yang, Vijay Prakash Dwivedi, Wei-Hsian Yin, Jeng Wei, Stefan Winkler \IEEEmembership{Fellow, IEEE}
\thanks{Submitted for review to ``IEEE Journal of Biomedical and Health Informatics'' on dd.mm.yyyy.}
\thanks{Corresponding author: V. Schlegel (viktor\_schlegel@asus.com).  V. Schlegel, A. R. Kashyap, T.-T. Nguyen, T.-H. Yang, V. P. Dwivedi, and S. Winkler are with ASUS Intelligent Cloud Services (AICS), Singapore.  W.-H. Yin and J. Wei are with Cheng Hsin General Hospital, Taipei.  S. Winkler is also with the National University of Singapore (NUS).}}
\begin{document}

\maketitle

\begin{abstract}
Computerised clinical coding approaches aim to automate the process of assigning a set of codes to medical records. While there is active research pushing the state of the art on clinical coding for hospitalized patients, the outpatient setting---where doctors tend to non-hospitalised patients---is overlooked. Although both settings can be formalised as a multi-label classification task, they present unique and distinct challenges, which raises the question of whether the success of inpatient clinical coding approaches translates to the outpatient setting. This paper is the first to investigate how well state-of-the-art deep learning-based clinical coding approaches work in the outpatient setting at hospital scale. To this end, we collect a large outpatient dataset comprising over 7 million notes documenting over half a million patients. We adapt four state-of-the-art clinical coding approaches to this setting and evaluate their potential to assist coders. We find evidence that clinical coding in outpatient settings can benefit from more innovations in popular inpatient coding benchmarks. A deeper analysis of the factors contributing to the success---amount and form of data and choice of document representation---reveals the presence of easy-to-solve examples, the coding of which can be completely automated with a low error rate.
\end{abstract}


\section{Introduction}
Medical records are primary sources of documentation of patient care, disease progression, and healthcare operations. To make these potentially unstructured records findable, accessible, and interoperable, they are codified by clinical coders according to a standardised vocabulary, such as the International Classification of Diseases (ICD) ontology~\cite{WHO1993ICD-10Diseases}---a hierarchically arranged vocabulary of standardised codes describing medical conditions, symptoms, diagnoses and hospital procedures. These codes, in turn, are used to claim reimbursement from medical insurance, to optimise resource allocations, or as a basis to select participants for clinical trials.  Being a fundamental building block for these operations, it is important to maximise the accuracy and efficiency of the coding process, which has given rise to computer-assisted clinical coding tools \cite{Stanfill2010ASystems}. Research in this direction promises to improve both the speed of clinical coding and the quality of resulting codes, easing the burden of clinicians and coders alike.

Despite steady progress in clinical coding of inpatient discharge summaries \cite{Mullenbach2018ExplainableText,Nguyen2023Mimic-IV-ICD:Classification,Kaur2023AI-basedReview,Nguyen2023ACoding}, there has been a lack of attention to other coding settings from the NLP community. One such example is the clinical coding of medical records that document patients’ ambulatory or \emph{outpatient} hospital visits. While the setting appears closely related to the inpatient setting, formalised using the same machine-learning task of multi-label document classification, clinical coding in the outpatient setting poses distinct challenges which call for a deeper investigation.

The first difference is the underlying document type and its purpose. While discharge summaries describe a patients’ course during their stay in a hospital, outpatient notes (similarly to inpatient progress notes) document a single doctor-patient encounter. 
Discharge summaries tend to be long and self-sufficient documents \cite{Wimsett2014ReviewReview}. Conversely, outpatient notes are much shorter; a collection of notes might document the progression of an ongoing condition. Consequently, outpatient notes often contain redundant information~\cite{Rule2021LengthCenter}. For example, when a patient visits a hospital to renew their medication without any change to the underlying condition, the doctor might copy different parts of existing documentation into the new note, including the notes’ text as well as the ICD codes.

\begin{figure*}[t]
\centering
\resizebox{0.97\textwidth}{!}{\includegraphics[clip, trim=4cm 1cm 0.5cm 8.4cm, width=1\textwidth, page=3]{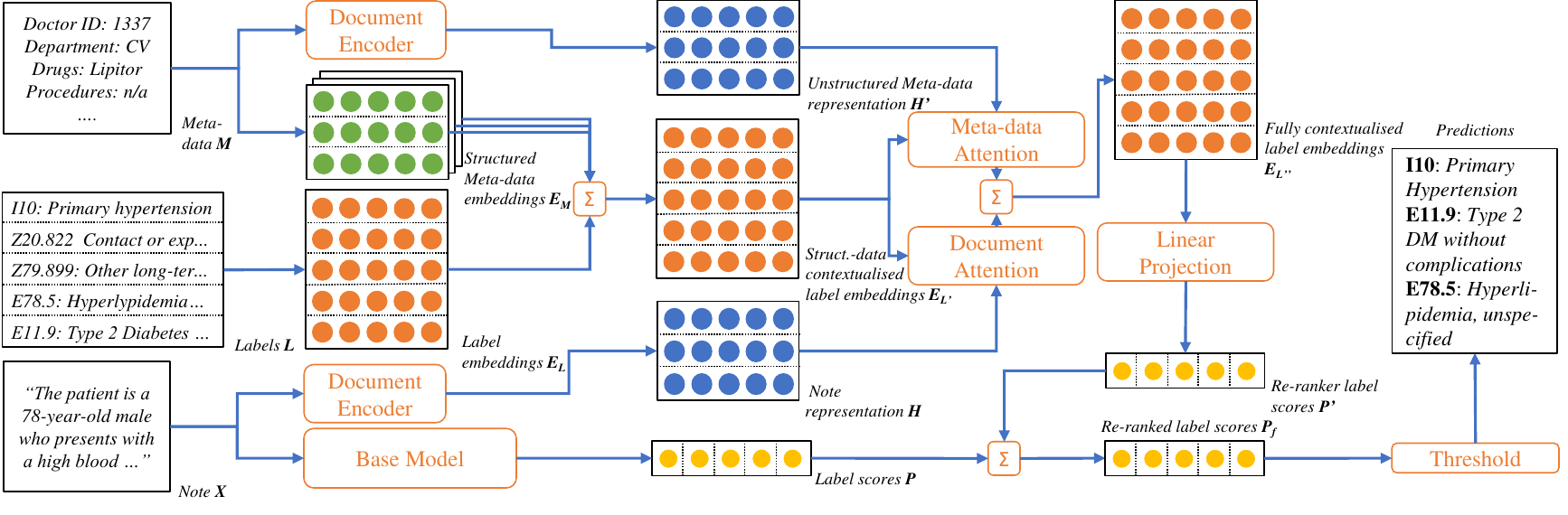}}
\caption{Overview of our proposed \textsc{OPD-Reranker} architecture which is optimised to re-rank the predictions of an optimised base model, taking into account available structured and unstructured additional (meta-)information in addition to the text contained in an outpatient note.}
\label{fig:arch}
\end{figure*}

Secondly, discharge summaries are typically written by doctors and later handed over to medical coders, i.e., different staff with different responsibilities \cite{Roberts2018ClinicalData}. As a result, discharge summaries must be self-contained, as they are the only way to convey information between summary author and coder~\cite{Nouraei2016AccuracyManagement}. This is not necessarily the case in outpatient settings, where doctors both document the visit and assign appropriate codes~\cite{Liang2020PhysicianCoding}. Therefore, information required to correctly codify a note might be omitted due to time constraints, as doctors might favour operational efficiency over the completeness of documentation~\cite{Schilling2010PerceivedEncounters}. For example, a doctor might omit important details in the note (e.g., whether a patient experienced pain in their left or right leg), and instead simply select the appropriate ICD code to provide additional context (e.g., \emph{``M79.662: Pain in left lower leg''}). 
This presents a major challenge to automated coding approaches because succeeding in the task requires the ability to infer such details, e.g., based on the prevalence of certain codes, by identifying distinct writing styles of doctors and learning their coding preferences~\cite{Pollard2013HowRecord} or by relying on additional available information, such as medications.

In this paper, we set out to investigate how well state-of-the-art automated clinical coding approaches are equipped to address the \emph{outpatient coding} task. To the best of our knowledge, this is the first study to investigate the feasibility of predicting ``billable'' (i.e., directly usable for reimbursement purposes) ICD10 codes. The investigation is carried out on a large-scale dataset comprised of more than seven million clinical notes describing outpatient encounters of more than 550k patients, contributed by more than 200 doctors from over 50 outpatient departments. Despite the differences outlined above, we find that advances in state-of-the-art inpatient clinical coding largely translate to the outpatient setting. We propose a flexible architecture to further improve performance by exploiting available structured and unstructured additional information. We further show that simple, data-oriented adaptations drastically reduce training time and improve training stability. Finally, we present a method to exploit model confidence on easy-to-classify examples to ``automate'' these with minimal false-positive rate. We conclude by making a set of recommendations for researchers and practitioners to support them in similar endeavours. 




\begin{table*}[t]
    \centering
    \caption{Statistics of the raw and processed OPD (``outpatient departments'') datasets in comparison with the most similar inpatient dataset, MIMIC-IV-ICD10.}
    \label{tab:dataset}
    \small
    \begin{tabular}{l ccc c ccc c ccc}
    & \multicolumn{3}{c}{\textbf{OPD-raw}} & & \multicolumn{3}{c}{\textbf{OPD-dedup}} & &  \multicolumn{3}{c}{\textbf{MIMIC-IV-ICD10}}\\
    & Train & Dev & Test & & Train & Dev & Test & & Train & Dev & Test \\
    \hline
    Document Type & \multicolumn{7}{c}{{Outpatient Notes}} & & \multicolumn{3}{c}{{Discharge Summaries}} \\
    Language & \multicolumn{7}{c}{{English and Chinese}} & & \multicolumn{3}{c}{{English}} \\
    Codes entered by & \multicolumn{7}{c}{{Doctor}} & & \multicolumn{3}{c}{{Clinical Coder}} \\
    Document Author is its Coder & \multicolumn{7}{c}{{\Checkmark}}  & & \multicolumn{3}{c}{{\XSolidBrush}} \\
    Code type & \multicolumn{7}{c}{{Diagnoses}} & & \multicolumn{3}{c}{{Diagnoses and Procedures}} \\
    \hdashline
    Number of Documents & 7,463K & 13,282 & 13,274 & & 2,381K & 4,323 & 4,378 & & 110,442 & 4,017 & 7,851\\
    Number of Patients & 554,917 & 1,000 & 1,108 & & 554,917 & 1,000 & 1,108 & & 59,114& 2,189 & 4,380 \\
    Number of Distinct Codes & 18701 & 1492 & 1494 & & 2588 & 1492 & 1494 & & 25,230 & 6,738 &  9,159 \\
    Mean Document Length (characters)  & 712 & 758 & 679 & & 396 & 380 & 368 & & 10,146 & 10,215 & 10,022 \\
    Mean \# Codes per Document & 2.85 & 2.86 & 2.78 & & 2.39 & 2.40 & 2.30 & & 16.1 & 16.2 & 15.8\\ 
    Distinct \% Codes Unseen in Train & - & 0.7\% & 0.5\% & & - & 15.8\% & 15.6\% & & - & 13.3\% & 6.4\% \\
    \hline
    \end{tabular}
\end{table*}


\section{Clinical Coding in Outpatient Settings}
In this section, we formulate the task of clinical coding and describe the corpus of outpatient notes and evaluation metrics used in the study.

\subsection{Task Formulation and Model Architectures}
\label{sec:models}
Clinical coding is formulated as document-level multi-label classification, where $n$ out of $N$ possible labels need to be assigned to an input document $\mathbf{X}$. Typically, $N$ is much larger than $n$, with an unbalanced label distribution~\cite{Johnson2016MIMIC-IIIDatabase,Johnson2023MIMIC-IVDataset}. 

To overcome these challenges associated with clinical coding, most deep learning-based techniques utilise two key elements: The first is a \emph{document encoder}, a neural network that combines the hidden representations of individual tokens to create a representation $\mathbf{H}$ of the input document $\mathbf{X}$. The second component is a \emph{label attention} mechanism which 
is used to obtain the label-specific document representation $\mathbf V$. These label-specific representations serve as input to the label classification layer to obtain probability vector $\mathbf{P}$ for each label. The entire architecture is trained end-to-end by minimising the binary cross-entropy loss between ground truth and predicted label probabilities. During inference, labels are selected as predictions based on a decision rule, such as the probability of a label being above a pre-defined threshold.

In our study, we focus on the following four approaches, in ascending order of their performance on MIMIC-IV-ICD10~\cite{Nguyen2023Mimic-IV-ICD:Classification}:
\subsubsection{CAML: Convolutional Attention Network for MultiLabel Classification}
The CAML architecture~\cite{Mullenbach2018ExplainableText} was the first to employ the label-attention mechanism to obtain label-specific document representations for each ICD code. It uses a single-layer CNN as the document encoder. 

\subsubsection{LAAT: Label Attention Model}
The LAAT model uses an LSTM network as document encoder \cite{Vu2020AText}. They improve the label attention layer with a structured self-attention mechanism that makes use of the hierarchy of the ICD ontology.

\subsubsection{MSMN: Multiple Synonyms Matching Network}
The MSMN architecture \cite{Yuan2022CodeCoding} enhances label attention by ICD code synonyms derived from the UMLS meta-thesaurus \cite{Bodenreider2004TheTerminology}. This is enabled by a multihead-synonym mechanism, which attends to synonyms to enhance the learned representation of the label embeddings.

\subsubsection{OPD-LM-LAAT}
This method employs pre-trained language models (PLM) \cite{Huang2022PLM-ICD:Models} as document encoders and the label attention layer of LAAT \cite{Vu2020AText}. Similarly, we investigate the application of a hospital-specialised language model.

\subsubsection{OPD-Reranker} Since textual records of outpatient encounters alone can be incomplete, we develop a simple architecture that takes into account available structured (e.g., lab results) and unstructured (e.g., imaging reports) information to re-rank the predictions of another (base) model (Figure~\ref{fig:arch}). Learnable embeddings $\mathbf{e}_m$ represent entries of each structured modality $m$ and a document encoder obtains the note representation $\mathbf{H}$ of unstructured information $\mathbf{H}'$. These are used obtain the final label scores in the following way:
\begin{align*}
    \mathbf{E}_{L'} &= \mathbf{E}_L \oplus \sum\nolimits_{m\in\mathcal{M}}\mathbf{e}_m \\
    \mathbf{E}_{L''} &= Attn_N(\mathbf{E}_L', \mathbf{H}, \mathbf{H}) + Attn_M(\mathbf{E}_L', \mathbf{H}', \mathbf{H}') \\
    \mathbf{P}' &= W_P \mathbf{E}_{L''} + b_P \\
    \mathbf{P}_f &= \mathbf{P}' + \mathbf{P}
\end{align*}
where $\mathbf{P}$ is the base model prediction, $\mathbf{E}_L$ are the ICD-label embeddings, $\mathcal{M}$ is the set of all available structured information modalities, $W_P$ and $b_P$ are the learnable weight matrix and bias vector of the final projection layer, respectively, and $Attn(Q,K,V)$ calculates the multi-head attention with query $Q$, key $K$ and value $V$ (and the corresponding learnable weights $W_Q$, $W_K$ and $W_V$ for each head and $W_O$ to project the concatenation of each heads' outputs). More specifically, $Attn_N$ attends to each token of the hidden note representation $\mathbf{H}$ and $Attn_M$ attends to each token of the representation of unstructured additional information $\mathbf{H}'$. $\oplus$~denotes label-wise addition, where the vector representing the sum of all modalities' $\mathbf{e}$ is added to each label embedding as row of the matrix $\mathbf{E}_L$.

\subsection{Evaluation Metrics}
Adoption of automated clinical coding tools can be challenging specifically in outpatient scenarios, as coders might feel ``hostile'' towards automated tools which could potentially replace them \cite{Stanfill2008CodingCoding}. Furthermore, state-of-the-art results on inpatient benchmarks \cite{Yang2022KnowledgeCoding,Nguyen2023Mimic-IV-ICD:Classification} have shown that the performance of existing automated clinical coding tools renders them unsuitable as replacement for human coders.

Therefore, in addition to the usual metrics employed for evaluation of multi-label classification problems, i.e., AUC and F1 (both micro- and macro-averaged), we also use metrics that emphasise the \emph{assistance} aspects of clinical coding~\cite{Campbell2020Computer-assistedProfessionals}. Therefore, we measure Recall@$k$, i.e., how many correct labels out of all correct ones are in the top $k$ scored predictions when ranked by their predicted probability. This allows us to approximate the performance in the scenario where a clinical coding system recommends a set of labels to a coder, leaving the final decision to a human. We set $k$ to 5, based on insights about the working memory capacity of humans \cite{Engle2004WhatCapacity}. Additionally, Recall@$k$ is \emph{threshold-free}, therefore, no decision rule is needed to convert from label probabilities to the predicted label set. This allows us to compare the predictions directly, without being influenced by their decision rules. 

In addition to the recall-based metric, to understand the performance on each evaluation sample, we calculate an \emph{instance-averaged} iF1 score---the harmonic mean between precision and recall---for each instance and take the mean across all instances. For the cases with iF1 of 1, the predictions are an exact match with the correct ground truth labels. 

\subsection{Corpus of Outpatient Notes}

To build a dataset to optimise data-driven clinical coding approaches, we collected outpatient encounter data from the Cheng Hsin General Hospital in Taipei, Taiwan, after obtaining approval from their institutional review board (number (774)109A-14). This dataset consists of over 7 million outpatient notes recording visits of over 550,000 patients, codified with their corresponding ICD10 labels. The statistics of the dataset are described in Table~\ref{tab:dataset}.

The input documents are shorter than their inpatient counterparts, fewer codes are assigned per note on average, and procedure codes are excluded from the documentation, which might suggest that the task is considerably easier. However, for outpatient encounters in Taiwan hospitals, attending doctors codify the documents \cite{Liang2020PhysicianCoding}, while coders spot-check these records a posteriori. This may lead to omitted textual information described earlier.  
Another challenge is bilinguality---while the primary documentation language is English, doctors may write parts of the note in (traditional) Chinese.  

Similar to MIMIC-IV-ICD10, codes are unevenly distributed, such that 50\% of the ICD10 codes are associated with five or fewer training instances (five for MIMIC-IV). Furthermore, the ratio of patients to medical records is much higher than comparable inpatient coding datasets (e.g., two discharge summaries on average per patient for MIMIC-IV). This is relevant, because unlike discharge summaries, which summarise a whole hospital course, outpatient notes keep record of each doctor-patient encounter. Many encounters are similar, e.g., when a patient presents with a chronic illness or needs to renew a drug prescription. To speed up the documentation process, doctors ``ditto'' or copy information from previous encounters, which may include free text, prescriptions and assigned ICD codes. As such, outpatient documentation more closely resembles inpatient progress notes, which describe the daily progress of a patients' treatment.

To circumvent leakage from training to evaluation data due to this ``ditto'' practice, we split the dataset into training and evaluation portions by patient. This ensures that models are evaluated on their capability to generalise to new cases, rather than memorising patients from training data. Specifically, we hold out 1000 and 1108 random patients for the development and test sets, respectively, to have a similar number of cases in both sets. To further align the evaluation with potential application scenarios, we remove ``ditto'' duplicates from evaluation data, since the labels for these cases are taken over from previous encounters, and there is no need for automated tools to predict them.  Because ``ditto'' information is not explicitly marked, we approximate duplicates by ordering records by date, matching ground truth ICD-10 codes for each patient and removing subsequent records where the label sets are an exact match. This leaves us with 4323 and 4378 notes for the development and test sets, respectively (Table~\ref{tab:dataset}).

\subsection{Data pre-processing}
While removing duplicates makes sense for evaluation data, it is unclear whether duplicate entries can help models during training. As we detect duplicates based on patient and label set, the input text might be changed between duplicates, due to e.g., incoming lab results or new developments of the same disease. Furthermore, while low-frequency codes have been reported as a challenge for clinical coding models~\cite{Yang2022KnowledgeCoding}, their practical impact on the problem is often unclear, even if many labels in the dataset are rare.

To answer these questions empirically, we optimise a baseline CAML model on 3 versions of the data: the full training set, a de-duplicated version, and a de-duplicated version where only labels that are observed at least 100 times in the training set are retained (\textsc{Full}, \textsc{Dedup} and \textsc{Min100}, respectively). Note that for the development and test sets, we do not remove any labels. We compare the performance among them and to the theoretical best performance, where we predict a label if it's in the ground truth and appears at least 100 times in the training set (\textsc{Oracle}). The results are reported in Table~\ref{tab:dedup-and-min}. Regarding duplicates, we see that the model greatly benefits from removing ``ditto'' with an absolute performance improvement of 14 points over the model that was trained on the full set, while reducing training time by 30\%. Disregarding rare labels during training has limited impact (below 3 points) for an oracle that predicts any other label correctly.  For actual models there is almost no noticeable performance degradation. The training time is further reduced, which stems from a combination of faster convergence and lower parameter count due to the reduced number of labels and their embeddings.

\begin{table}[!tbh]
\small
\caption{R@5 on the development set of CAML-models trained in different data configurations.}
\label{tab:dedup-and-min}
\begin{tabular}{lccc}
\textbf{Category} & \textbf{R@5 dev} & \textbf{Train time (h)}\\
\hline
\textsc{Oracle} & $99.26$ & - \\
\textsc{Oracle+Min100} & $96.29$ & - \\
\hdashline
\textsc{Full} & 55.66 & 30.2h \\
\textsc{Dedup} & 69.06 & 19.0h \\
\textsc{Dedup+Min100} & 68.41 & 9.4h \\
\end{tabular}
\centering
\end{table}

Based on these insights, we continue our experiments with a ``ditto''-deduplicated training set where all labels occurring less than 100 times are removed.
The statistics of our training set are described in Table~\ref{tab:dataset}, in the column ``OPD-dedup''. Notably, the average length of non-ditto instances is considerably shorter (396 vs 712). This suggests that the medical records of patients who visit the hospital repeatedly for treatment of the same condition ``grow'' by incorporating new information, such as lab results, new symptoms or similar.








\section{Empirical Study}
We broadly aim to investigate the feasibility of automating clinical coding for outpatient departments. More specifically, we seek evidence towards the following research questions:
\begin{enumerate}[(i)]
    \item Can solutions to automated coding of inpatient discharge summaries be applied to outpatient clinical coding? Do improvements upon the state-of-the-art in inpatient coding translate to outpatient settings?
    \item What is the impact of different document encoders on coding performance?
    \item What is the relation between training data and performance? Does more training data translate to better performance?
    \item On what proportion of the data do models' predictions exactly match the set of ground-truth labels? Can these examples be identified reliably?
\end{enumerate}

\begin{table*}[!t]
\centering
\caption{AUC, F1 and Recall@$5$ scores for the baselines on the development and test sets of the OPD dataset. 
}\label{tab:overall}

\resizebox{1.85\columnwidth}{!}{
\begin{tabular}{l cc:ccc:c}
\multirow{2}{*}{\bfseries Model} &\multicolumn{2}{c}{\bfseries AUC} &\multicolumn{3}{c}{\bfseries F1} & \textbf{Recall} \\
& Macro & Micro & Macro & Micro & Instance & R@5 \\
\hline
CAML & $88.67$/$88.84$ & $96.16$/$96.54$ & $14.93$/$14.96$ & $42.12$/$41.79$ & $51.58$/$51.43$ & $68.41$/$68.81$\\
LAAT  & $94.83$/$94.98$ & $98.26$/$98.43$ & $17.19$/$17.31$ & $58.43$/$57.56$ & $59.38$/$58.51$ & $74.83$/$75.25$ \\
MSMN & $\mathbf{97.40}$/$\mathbf{97.98}$ & $\mathbf{98.67}$/$\mathbf{99.01}$ & $16.92$/$17.21$ & $57.34$/$57.43$ & $59.19$/$59.58$ &  $74.23$/$75.24$\\
OPD-LM-LAAT  & $94.98$/$95.70$ & $98.58$/$98.82$ &  $20.44$/$21.21$ & $62.37$/$62.52$ & $66.81$/$66.90$ & $79.12$/$79.62$\\
\hdashline
\textsc{OPD-Reranker}  & $94.58$/$95.14$ & $98.33$/$98.54$ &  $\mathbf{20.83}$/$\mathbf{21.46}$ & $\mathbf{62.94}$/$\mathbf{63.09}$ & $\mathbf{67.05}$/$\mathbf{67.47}$ & $\mathbf{79.23}$/$\mathbf{80.06}$\\
\end{tabular}
}
\end{table*}

Regarding \emph{(i)}, it is unclear if discharge summary coding approaches will perform well in the outpatient scenario due to the difference between in- and out-patient data. To this end, we adapt the state-of-the-art clinical coding approaches discussed in Section~\ref{sec:models}\cite{Mullenbach2018ExplainableText,Vu2020AText,Yuan2022CodeCoding,Huang2022PLM-ICD:Models} that have been shown to perform well in inpatient settings \cite{Nguyen2023Mimic-IV-ICD:Classification} to the outpatient scenario. 

For question \emph{(ii)}, there is conflicting evidence whether using transformer-based language models as document encoder improves upon traditional word embeddings~\cite{Yuan2022CodeCoding,Huang2022PLM-ICD:Models}, and if their domain-specific pre-training is beneficial. We perform an ablation study to observe the difference in performance when using randomly initialised, domain-specific, or hospital-specific embeddings, for both word-vector and language-model based document encoders. 

Answering question \emph{(iii)}, investigating model performance as function of the training set size can provide insights on obtaining well-performing models when constrained by available training data or hardware resources. 

Finally, for question \emph{(iv)}, we threshold over model prediction probabilities to identify instances where predictions exactly match the ground truth label, subject to a fixed false positive rate. In practice, examples identified in such a way can be exempted from human review, further reducing the cognitive load of coding doctors.

\subsection{Implementation Details}
We adapt all evaluated approaches to the ICD-10 setting. For models that rely on word embeddings, we optimise these on available training data in line with literature \cite{Mullenbach2018ExplainableText}. For OPD-LM-LAAT that uses a language model, we train a custom BERT MLM with three layers, four heads, hidden dimension of 512 and feed-forward dimension 2048 on all training notes. We decide to train from scratch, to accommodate tokenisation of domain-specific terms \cite{Lehman2023DoModels}. 

For the MSMN model, we obtain code descriptions from the 2016 release of the ICD10-CM ontology\footnote{ \url{https://www.cms.gov/Medicare/Coding/ICD10/2016-ICD-10-CM-and-GEMs}} and their synonyms from the UMLS Metathesaurus~\cite{Bodenreider2004TheTerminology}.

For the re-ranker model, we utilise prescribed medication, carried out procedures, doctor ID and the doctor's department as structured information, and the textual description (i.e., the name) of medications and procedures as unstructured information. Note that because multiple prescriptions and procedures can be associated with a single outpatient visit, we take the average of their embeddings (multi-hot-encoding). We use OPD-LM-LAAT as the base model to re-rank, freeze its weights when obtaining $\mathbf{P}$ and $H$ and share its document encoder with the reranker. 

We train all models according to the hyper-parameters reported in the respective literature and use early stopping for CAML, LAAT and MSMN and monitor Recall@$5$. Further hyper-parameters are reported in the Appendix. All experiments were carried out on a NC24sv3 Azure instance with four V100 GPUs, 24 vCPUs and 448GB of RAM. CAML and LAAT were trained on a single GPU, while OPD-LM-LAAT and MSMN were trained using two GPUs each.

\subsection{Results and Analysis}


In this section, we report the results of our study and relate them to the guiding questions.
\subsubsection{Existing clinical coding architectures can be applied in the outpatient context and additional information is helpful} Table~\ref{tab:overall} reports the performance of all four models on the development and test sets. The general performance trend observed on inpatient benchmarks carries over to our OPD dataset as well. The MSMN model constitutes an exception, as it performs worse than LAAT in terms of R@5 score but better regarding the AUC metrics. One possible reason might be that---as reported by others \cite{Nguyen2023Mimic-IV-ICD:Classification}---MSMN tends to perform better on rare codes, which are excluded from our training set by design. Incorporating additional information by training our proposed re-ranker model on top of the best performing OPD-LM-LAAT further improves performance. 

\begin{figure}[bh]
\centering
\resizebox{\columnwidth}{!}{\includegraphics[clip, trim=0cm 1cm 0cm 0cm, width=1\columnwidth]{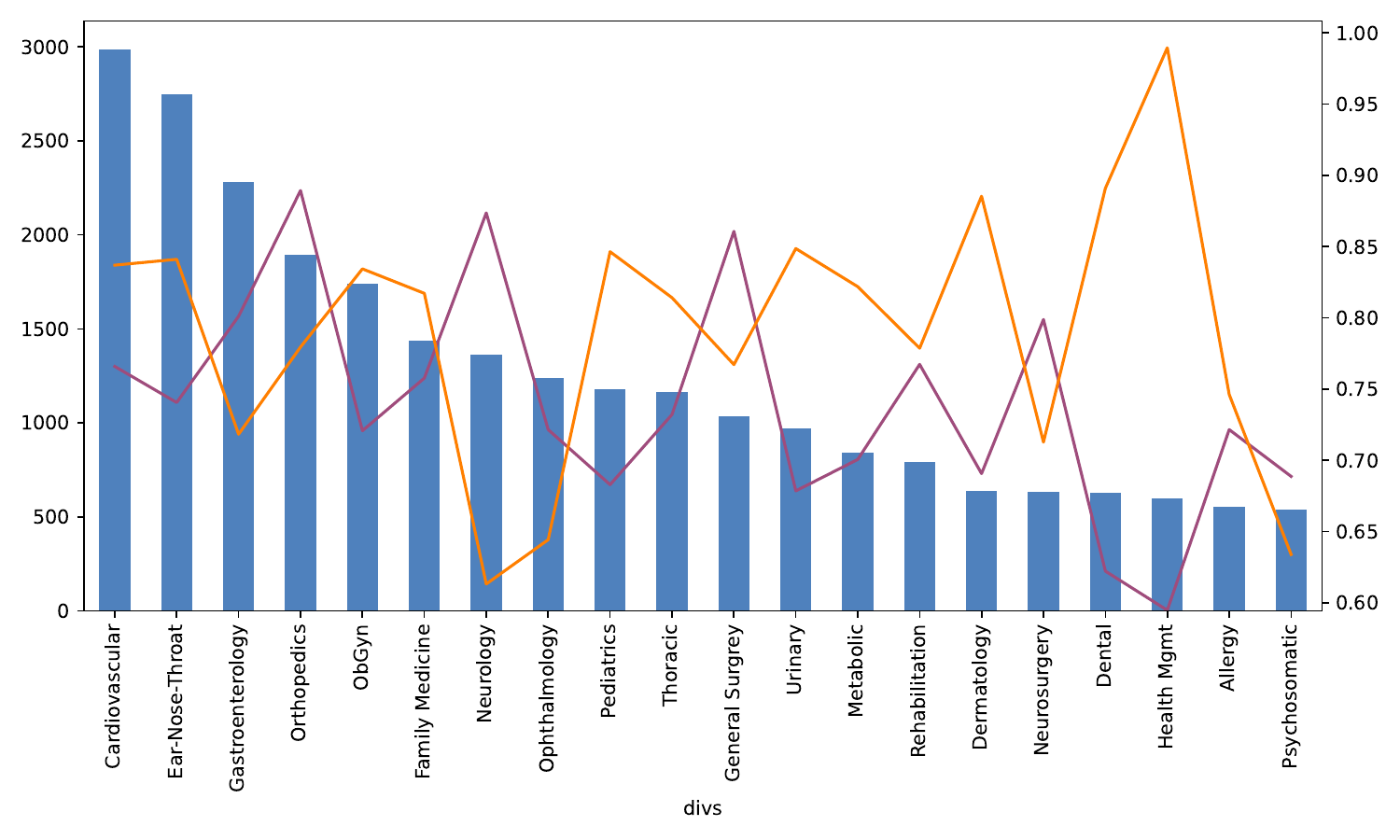}}
\caption{Performance breakdown for the 20 most contributing departments, ranked by frequency of encounter ({\color{blue}blue} bar, left scale, 10x). {\color{orange}Orange} plot represents R@5 score (right scale); {\color{purple}Purple} plot represents the average number of distinct labels observed in a department per year (left scale).}
\label{fig:by-dept}
\end{figure}

\begin{figure}[bh]
\centering
\resizebox{\columnwidth}{!}{\includegraphics[clip, trim=5cm 1cm 4cm 0cm, width=1\columnwidth]{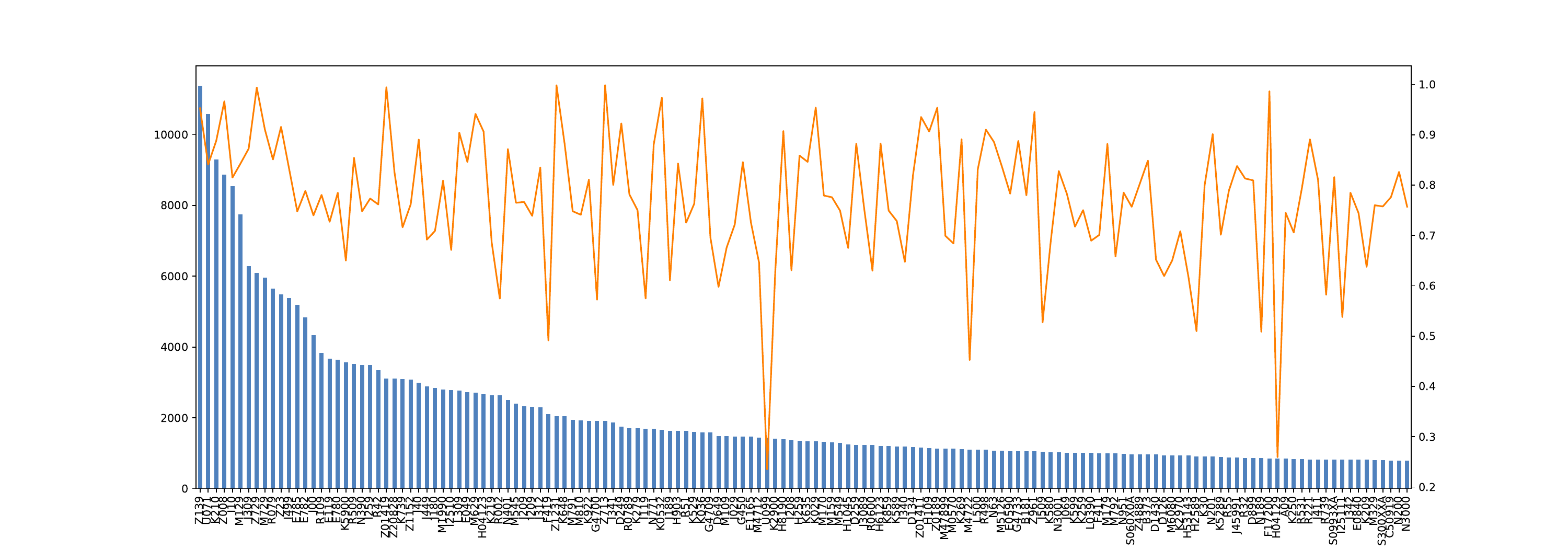}}
\caption{Performance breakdown for the 150 most frequent labels, ranked by label frequency ({\color{blue}blue} bar, left scale). {\color{orange}Orange} plot represents R@5 score (right scale).}
\label{fig:by-label}
\end{figure}

Even after removal of ``ditto'' instances, patients can appear more than once in the test set, if they present with new conditions and therefore new codes. Comparing the performance on such ``recurring'' patients to patients at their first visit, the performance drops (Recall@$5$ of $82.63$ vs $77.20$ for OPD-LM-LAAT), possibly because new codes are added with relatively little accompanying documentation~\cite{Schilling2010PerceivedEncounters} and because the training data is biased towards ``first-visit'' patients due to our de-duplication method.

Further inspection of the performance broken down by department (Figure~\ref{fig:by-dept}), reveals that the performance varies greatly by department, with no correlation between a departments' score and its number of examples in the training set. There is, however, a strong (anti-)correlation with the overall number of labels encountered per year (Spearman's $r=-0.68$, $p<.005$). Similarly, Figure~\ref{fig:by-label} shows that the performance on 150 most frequent labels (which constitute 55\% of the overall test set), correlates only weakly with their support in the training set (Spearman's $r=0.22$, $p<.005$).


\subsubsection{Domain-specific pre-training is helpful} We investigate the impact of the document encoder choice by fixing the label encoding mechanism and comparing different document encoders. We use LAAT for word-vector based document encoders  and OPD-LM-LAAT for language-model based document encoders, as they both implement the same label attention mechanism \cite{Huang2022PLM-ICD:Models}. We additionally use LAAT's word embeddings\cite{Vu2020AText} and BioLM, a RoBERTa-base model optimised on MIMIC-III notes \cite{Lewis2020PretrainedState-of-the-Art} as domain-specific document encoders (labelled mimic). We also report performance on randomly initialised word embeddings and LM weights.
Table~\ref{tab:doc-enc} shows that LM-based document encoders outperform word-vector based ones.

\begin{table}[hb]
\centering
\caption{iF1 and R@5 scores on the development and test datasets when using word vector and Language Model based encoders. 
Each category is optimised on hospital data, initialised randomly or optimised on MIMIC-III notes, respectively.}
\label{tab:doc-enc}
\begin{adjustbox}{max width=\columnwidth}
\begin{tabular}{lcccc}
\textbf{Note Encoder}  &\multicolumn{2}{c}{\textbf{iF1}} &\multicolumn{2}{c}{\textbf{R@5}} \\
& dev & test & dev & test \\
\hline
OPD-w2v & $59.38$ & $58.51$ & $74.83$ & $75.25$\\
random-w2v & $57.92$ & $57.53$ & $71.36$ & $72.54$ \\
mimic-w2v & $57.26$ & $55.28$ & $73.01$ & $72.97$ \\
\hdashline
OPD-LM & $\mathbf{66.81}$ & $\mathbf{66.89}$ & $\mathbf{79.12}$ & $\mathbf{79.62}$ \\
random-LM &  $5.00$ & $5.12$ & $16.83$ & $16.45$ \\
mimic-LM & $66.62$ & $66.64$ & $78.45$ & $78.75$ \\
\end{tabular}
\end{adjustbox}

\end{table}

Furthermore, domain-specific pre-training improves performance compared to random embeddings and a randomly-initialised language model. The latter makes sense, as language models are typically trained for millions of steps to converge~\cite{Devlin2018}. Finally, comparing the performance of domain-specific to hospital-specific document encoders, we see a clear benefit for both word embeddings and hospital-specific languages, as they outperform their MIMIC-III counter-parts. This is especially remarkable because the hospital-specific model is much smaller compared to \mbox{BioLM} (24.5 vs.\ 124.4 million parameters respectively).



\subsubsection{Models can achieve good performance with little training data} 
\begin{figure}[thb]
\begin{center}
\begin{tikzpicture}
\begin{axis}[
   title={},
   width=0.95\columnwidth,
   height=0.65\columnwidth,
   legend pos=south east,
   legend columns=2, 
   ymajorgrids=true,
   xlabel=\% train data,
   ylabel=\% best score,
   grid style=dashed,
   yticklabel={\pgfmathparse{\tick*100}\pgfmathprintnumber{\pgfmathresult}\%},
]
\addplot [
    color=blue,
    mark=triangle,
    legend entry=iF1 test
]
    coordinates {(0.01, 0.71) (0.05, 0.88) (0.1, 0.92) (0.25, 0.97) (0.33, 0.97) (0.5, 0.99) (0.75, 0.99) (0.90, 1.00)};
\addplot [   
    color=orange,
    mark=square,
    legend entry=R@5 test,
]
coordinates {(0.01, 0.76) (0.05, 0.90) (0.1, 0.94) (0.25, 0.97) (0.33, 0.98) (0.5, 0.99) (0.75, 0.99) (0.90, 0.99)};
\end{axis}
\end{tikzpicture}
\end{center}
\caption{Recall@5 and iF1 performance of OPD-LM-LAAT, when optimised on a random sample of 1, 5, 10, 25, 33, 50, 75, 90\% of data as fraction of score when optimised on full dataset.}
\label{fig:by-data}
\end{figure}
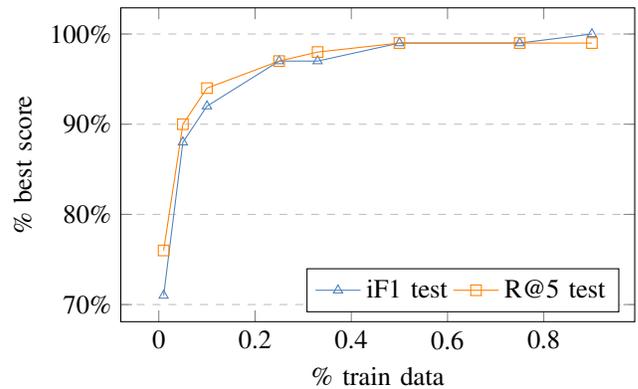

Figure~\ref{fig:by-data} shows the scores of OPD-LM-LAAT, when trained on random fractions of available training data. Concerning Recall@$5$, the scores saturate quickly, with a model trained on 5 percent of the data achieving 90 percent of the performance of the model trained on the full dataset. As such, more training data only yields diminishing returns of score improvement. The picture is less obvious for iF1 scores, where 10\% data is needed to surpass the 90\% performance threshold. Training a model on 50\% of the data can reach 99\% performance of the model trained on the full dataset for both Recall@$5$ and iF1.  

\begin{figure}[htb]
\centering
    \begin{tikzpicture}
    \begin{axis}[
        ybar=-10pt,
        x tick label style={font=\scriptsize\itshape},
        y tick label style={rotate=90, font=\scriptsize\itshape},
        height=11em,
        ylabel=\emph{\# instances},
        xlabel=\emph{score bins},
        width  = \columnwidth,
        bar width=4.5pt,
        ymajorgrids=true,
        y grid style=dashed,
        major y tick style = transparent,
        xtick={0,0.1,0.2,0.3,0.4,0.5,0.6,0.7,0.8,0.9,1.0},
        xmajorgrids=true,
        xminorgrids=true,
        legend pos=north west,
        legend cell align={left},
        legend columns=2,
        xmin=-0.05, xmax=1.05,
    ]
    \addplot[legend entry=\textsc{iF1}, color=blue, fill=blue]  coordinates {(0.05, 907) (0.15, 0) (0.25, 102) (0.35, 103) (0.45, 196) (0.55, 484) (0.65, 692) (0.75, 91) (0.85, 338) (0.95, 1465)};
    \addplot[legend entry=\textsc{Recall@5}, color=orange, fill=orange]  coordinates {(0.05, 318) (0.15, 11) (0.25, 87) (0.35, 158) (0.45, 43) (0.55, 529) (0.65, 301) (0.75, 157) (0.85, 85) (0.95, 2689)};
    \end{axis}
    \end{tikzpicture}
    
    \caption{Distribution of iF1 and Recall@$5$ scores on the test set of OPD-LM-LAAT.}
    \label{fig:f1-dist}
\end{figure}
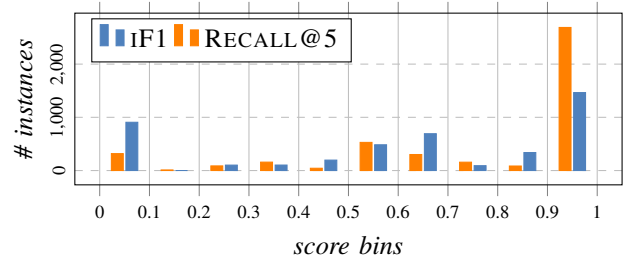

This quick saturation could hint at the presence of ``easy'' examples which the models quickly learn to solve correctly, achieving an iF1 and Recall@5 score of 1 on these. Indeed, Figure~\ref{fig:f1-dist} shows that the distribution of iF1 and Recall@5 scores is skewed towards the last $(0.9, 1.0]$ bin. In fact, $33.6\%$ and $61.4\%$ of the examples have an instance-averaged iF1 and Recall@$5$ score of 1, respectively. 

We find that overall, shorter examples with less codes tend to be easier (Spearman $r=-0.25$ and $r=-0.14$ correlation between iF1 score and length and number of codes, respectively, both $p<0.005$).

\subsubsection{Easy examples can be identified with low false-positive rate} 
Given an input note and the model predictions upon it, we are looking for a rule to decide whether the model predictions are an exact match to the ground truth label set. We evaluate the efficacy of the decision rule by measuring the number of correctly identified examples at maximum allowed false positive (max FP) rate of 5, 10, 15 and 20\%, respectively.

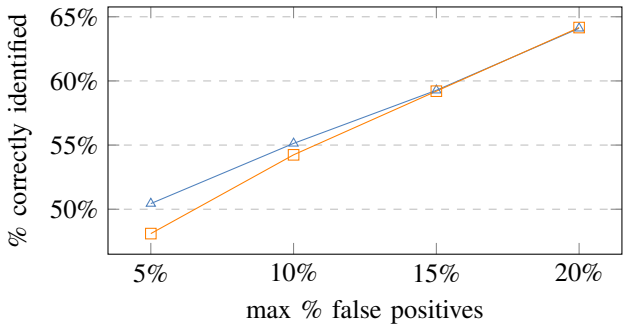
\begin{figure}[!bh]
\begin{center}
\begin{tikzpicture}
\begin{axis}[
   title={},
   width=0.95\columnwidth,
   height=0.55\columnwidth,
   xlabel=max \% false positives,
   ylabel=\% correctly identified,
   ymajorgrids=true,
   grid style=dashed,
   xticklabel={\pgfmathparse{\tick*100}\pgfmathprintnumber{\pgfmathresult}\%},
   yticklabel={\pgfmathparse{\tick}\pgfmathprintnumber{\pgfmathresult}\%},
]
\addplot [   
    color=blue,
    mark=triangle,
] coordinates {(0.05, 50.44) (0.1, 55.13) (0.15, 59.27) (0.20, 64.10)};
\addplot [   
    color=orange,
    mark=square,
] coordinates {(0.05, 48.10) (0.1, 54.24) (0.15, 59.20) (0.20, 64.16)};

\end{axis}
\end{tikzpicture}
\end{center}
\caption{Percentage of correctly identified instances with iF1 scores of 1 (out of all possible) by using confidence thresholding {\color{blue}with} and {\color{orange}without} calibration, subject to maximum false-positive rate, i.e., the number of examples identified incorrectly as having iF1 of 1 divided by all examples identified.}
\label{fig:conf-threshold}
\end{figure}

The decision rule can be chosen arbitrarily---for simplicity we use the confidence thresholding method: we choose an example if the predicted label probabilities are all above threshold $t_u$ and all other labels' probabilities are below $t_l$. Using the predictions of OPD-LM-LAAT, we exhaustively search all possible $t_u, t_l \in [0, 1]^2$ combinations in $0.05$ increments and select the one with the highest number of selected exact match predictions on the development set, subject to max FP. The resulting number of identified examples in the test set is shown in Figure~\ref{fig:conf-threshold}. Furthermore, previous research has shown that neural networks' probabilities do not necessarily represent the models' confidence~\cite{Vaicenavicius2019EvaluatingClassification}. To this end, we calibrate 
the models' predictions using label-wise isotonic regression~\cite{Zadrozny2002TransformingEstimates}, where for each label, all predictions on the development set are sorted into bins for which a piece-wise constant function is fit on ground-truth labels to map from raw to calibrated probabilities.

We find that this method indeed reduces the expected calibration error (ECE) for 73\% of the labels ($8.3 \cdot 10^{-4}$ to $7.6 \cdot 10^{-4}$, averaged across all dev/test labels), resulting in more faithful prediction probabilities. However it lowers the overall iF1 and R@5 scores. Nonetheless, the calibration seemingly helps to identify high- and low-confidence predictions better, as it improves the confidence thresholding method specifically for high $t_u=0.95$ and low $t_l=0.10$, i.e., at the lowest false positive error rate of $0.05$, as shown in Figure~\ref{fig:conf-threshold}. Overall, the best method can identify 50\% of the exact match instances of the best models (17\% of all test set notes) at the lowest max FP rate.

\section{Discussion and Related Work}
We relate our findings to the following three directions emerging in the literature on clinical coding: incorporating additional knowledge by improving label and document representations, improving performance on rare labels, and replicating performance on data other than MIMIC-III.

\subsubsection{Improving Representations} Different neural architectures have been proposed to represent \emph{input documents}, including CNNs~\cite{Mullenbach2018ExplainableText}, LSTMs~\cite{Vu2020AText} and pretrained language models \cite{Pascual2021TowardsOpportunities,Huang2022PLM-ICD:Models} or their combinations \cite{Zhou2021AutomaticMechanism}. For inpatient settings, conflicting claims regarding the efficacy of pretrained language models have been put forward~\cite{Yuan2022CodeCoding,Huang2022PLM-ICD:Models}, with the best-performing model on MIMIC-III employing a combination of both, the LSTM-based MSMN architecture to obtain an initial ranked list, followed by a transformer-based language model used for reranking~\cite{Yang2022KnowledgeCoding}. Regarding label representations, previous works have proposed the label-attention mechanism~\cite{Xie2018ACoding,Mullenbach2018ExplainableText}, which was further refined to incorporate label hierarchy and co-occurrences~\cite{Falis2019OntologicalText,Vu2020AText,Cao2020HyperCore:Coding}, label descriptions~\cite{Feucht2021Description-basedClassification} and synonyms~\cite{Yuan2022CodeCoding} and other external knowledge~\cite{Wang2022AKnowledge}.

We find that for outpatient coding, transformer-based document encoders clearly out-perform word embedding based ones. Furthermore, leveraging hospital-specific information by using a language model that was trained on hospital data further improves performance. Additionally, transformer-based encoders might perform so well in our setting, because the input documents are shorter than the 512 token limit of most transformers, such that no advanced input chunking strategies are required~\cite{Huang2022PLM-ICD:Models}.

\subsubsection{Rare labels} Some previous works made efforts to improve the coding frequency on rare~\cite{Yang2022KnowledgeCoding} or unseen labels~\cite{Rios2018Few-ShotSpaces,Lu2020Multi-labelGraphs}. Conversely, we find that from a practical perspective, rare and unseen codes hardly play a role in the overall performance of the models, when focusing on instance-averaged metrics, both in theory and practically.

\subsubsection{Replicating performance} Finally, some works have extended the evaluation of their methods beyond the MIMIC-III dataset. Most recently, MIMIC-IV benchmarks were introduced~\cite{Nguyen2023Mimic-IV-ICD:Classification,Edin2023AutomatedStudy}, including portions of records associated with ICD-10 codes, the up-to-date ontology used in practice. Others have applied their methods to other datasets, such as  data from other US hospitals~\cite{Zhang2020BERT-XML:Pretraining,Kavuluru2015AnRecords,Rios2019NeuralEMRs} or countries~\cite{Lin2017ArtificialNotes,Moons2020ARecords,Mayya2021Multi-channelSummaries,Dong2021ExplainableInitialisation}. To the best of our knowledge, all these studies focus on inpatient or emergency departments, as opposed to outpatient visits discussed in our paper.

\subsubsection{Label Quality}
\label{sec:qual}
Potentially erroneous human-assigned labels pose a well-documented problem for evaluation of automated clinical coding approaches~\cite{Horsky2017AccuracyVisits,Yeoh1993ClinicalOn.,Heywood2016ImprovingKey}. We also note that the performance of evaluated models plateaus around 80\% Recall@$5$, i.e. retrieving 4 out of 5 codes on average.  
To estimate the quality of annotations, we measure the (in)consistency of the labels by cross-referencing the ICD10 labels assigned to the same patient between inpatient and subsequent outpatient visits occurring within at most seven days. 
We find instances of code \underline{\smash{inconsistency}}, such as \textit{``E78.2: \underline{\smash{Mixed}} hyperlipidemia''} (inpatient) and \textit{``E78.5: Hyperlipidemia, \underline{\smash{unspecified}}''} (outpatient) for the same patient. A similar situation can be observed with \textit{``I70.203: Unspecified atherosclerosis of native arteries of extremities, \underline{\smash{bilateral legs}}''}, and \textit{``I70.202: Unspecified atherosclerosis of native arteries of extremities, \underline{\smash{left leg}}''}, for the two respective visits by a patient. Of 493 matches between the labels in our data, we find an inconsistency rate of 20.08\% at the level 3 or higher (i.e., two matched codes have the same chapter---the first three characters---but differ in any character afterwards) in the ICD10 codes. We deem these co-occurrences in quick succession unlikely (e.g., it is highly unlikely that the condition of mixed hyperlipedemia changes to another, unspecified hyperlipidemia within less than a week) and interpret them as coding errors. Extrapolating this finding to the full dataset, we assume that 80\% of our labels are correct, close to the best-performing re-ranker model Recall@$5$ score, suggesting that this model already reaches the possible performance ceiling. Note that this assumption is based on suggestive rather than conclusive evidence, as the selection criterion of patients that were transferred from inpatient to outpatient settings within a week might introduce an unknown bias.

\subsubsection{Study Limitations}
Similar considerations as with the \mbox{MIMIC-III and -IV} datasets apply: our study was carried out on a dataset from one hospital in a specific country. As such, even though encompassing over half a million patients, our dataset might lack diversity. To alleviate this and make our conclusions more robust, our findings will be replicated on other outpatient clinical coding datasets as future work. The only comparable study of outpatient clinical coding~\cite{Kuo2023ApplyingRecords} does not encompass the whole hospital and focusses only on five outpatient departments. More importantly, they only predict ``parent'' ICD-10 codes up to the first three characters (i.e., ``E11'' instead of the billable ``E11.9'' code). As such, their developed models are not directly able to predict billable codes; therefore the study is of preliminary nature. However, regarding by-department breakdown, they report trends similar to our results, both regarding (relative) per-department performance as well as the diversity of codes encountered per department.

Furthermore, we empirically choose the best hyper-parameters for each of the evaluated models. To obtain more robust final performance numbers, this process can be complemented by an exhaustive hyper-parameter search. Hyper-parameters can greatly impact the final performance \cite{Liu2021ParameterIt}, with the CAML model outperforming multiple subsequent incremental improvements after a careful selection of hyper-parameters. However, it has been shown to not out-perform the baseline model architectures selected for our study \cite{Nguyen2023Mimic-IV-ICD:Classification}, which gives us confidence that the trends in performance we report are robust to hyper-parameter choice.



\section{Conclusion}
In this paper we have investigated the feasibility of automated clinical coding approaches when applied to assisting doctors in the outpatient settings. Our results indicate that generally, advancements in the state-of-the-art on publicly available benchmarks for clinical coding~\cite{Johnson2016MIMIC-IIIDatabase,Johnson2023MIMIC-IVDataset,Nguyen2023Mimic-IV-ICD:Classification} can be transferred to the inpatient setting.

Based on our analysis, we formulate the following recommendations for researchers or practitioners who aim to replicate our results on other outpatient datasets. \textbf{Train your own Language Model}, as hospital-specialised, transformer-based document encoders have outperformed other document encoder approaches in our experiments.
\textbf{Remove noise}, such as duplicate entries or low-frequency codes, as it greatly improves training stability and speed. However, \textbf{utilise additional information}---our proposed re-ranking framework incorporates both structured and free-text additional information, which, as our experiments show, further improves performance over text-only approaches.
\textbf{Start training early}, even if lacking annotated data, as we have could achieve good performance with only a fraction of the available data.
\textbf{Find easy examples}, as labels for these can be pre-selected automatically, further increasing the operational efficiency of outpatient doctors.

To address the suspected label inconsistencies, we aim to further expand on our methods to detect them, for example by using model confidence scores or model interpretability methods to identify conflicting label assignments. This can be used to notify practitioners during the coding process, ultimately improving coding consistency and reducing errors. 






\section*{References}
\bibliography{main}
\bibliographystyle{ieeetr}

\appendix
\section{Further experiment details}
Regarding the choice of hyper-parameters, we leave all settings of all implementations as recommended by the authors of the corresponding papers, with the following exceptions:
    For CAML and LAAT, we train our own word embeddings on the training portion of our dataset using the code provided by CAML;
    For MSMN, we also obtain a word frequency list from our data and set the batch size to 8;
    For the OPD-LM-LAAT model, we implement a custom LAAT-based multilabel classifier in the PLM-ICD implementation \cite{Huang2022PLM-ICD:Models} to fit our LM architecture. We change the number of training epochs from 20 to 8, due to dataset size and train with batch size of 32; For the re-ranker model, we use two heads for both multi-head attention mechanisms. The dimensionality of all embeddings and hidden representations is 512. We train the re-ranker model for 5 epochs with batch size of 32.
For all models, we set the maximum input length to 512 tokens according to their respective tokenisation methods.
Comparison with the MIMIC-IV scores are reported below. Across MIMIC-IV and OPD, PLM-ICD should be compared to our OPD implementation OPD-LM-LAAT.

\begin{tabular}{lrrrrrr}
& \multicolumn{2}{c}{\bfseries MIMIC-IV}& &\multicolumn{2}{c}{\bfseries OPD} \\
\multirow{2}{*}{Model} &\multicolumn{2}{c}{F1} & & \multicolumn{2}{c}{F1} &  \\\cmidrule{2-3}\cmidrule{5-6}
&Macro &Micro & &Macro &Micro  & \\
\hline
CAML  & 4.61 & 53.32  & & 14.23 & 40.79 & \\
LAAT  &4.47 &55.40 & & 17.31 & 57.56  \\
MSMN  & 5.42 & 55.91 &  & 17.21 & 57.43  \\
PLM-ICD  & 4.90 & 56.95 & & - & - & \\
OPD-LM-LAAT  & - & - & & 21.21 & 62.52 & \\
\end{tabular}
\label{tab:opd-vs-mimic}

\end{document}